\documentclass[runningheads]{llncs}
\usepackage{adjustbox}
\usepackage{graphicx}
\usepackage{amsmath}
\usepackage{amsfonts}
\usepackage{float}
\usepackage{multirow}
\usepackage{threeparttable}
\usepackage{caption}
\usepackage{subcaption}
\usepackage{color}
\usepackage{url}
\makeatletter
\def\UrlAlphabet{%
      \do\a\do\b\do\c\do\d\do\e\do\f\do\g\do\h\do\i\do\j%
      \do\k\do\l\do\m\do\n\do\o\do\p\do\q\do\r\do\s\do\t%
      \do\u\do\v\do\w\do\x\do\y\do\z\do\A\do\B\do\C\do\D%
      \do\E\do\F\do\G\do\H\do\I\do\J\do\K\do\L\do\M\do\N%
      \do\O\do\P\do\Q\do\R\do\S\do\T\do\U\do\V\do\W\do\X%
      \do\Y\do\Z}
\def\UrlDigits{\do\1\do\2\do\3\do\4\do\5\do\6\do\7\do\8\do\9\do\0}
\g@addto@macro{\UrlBreaks}{\UrlOrds}
\g@addto@macro{\UrlBreaks}{\UrlAlphabet}
\g@addto@macro{\UrlBreaks}{\UrlDigits}
\makeatother
\newcommand*\samethanks[1][\value{footnote}]{\footnotemark[#1]}
\begin{document}
\title{SiamParseNet: Joint Body Parsing and Label Propagation in Infant Movement Videos}
\titlerunning{SiamParseNet}
%
\author{Haomiao Ni\inst{1}\thanks{These authors contributed equally to this work.} 
\and
Yuan Xue\inst{1}\samethanks 
\and
Qian Zhang\inst{2}
\and
Xiaolei Huang\inst{1}
}

\authorrunning{H. Ni \textit{et al.}}
\institute{College of Information Sciences and Technology, The Pennsylvania State University, University Park, PA, USA 
\and
School of Information and Control Engineering, Xi’an University of Architecture and Technology,
Xi’an, China
}
\maketitle              

\begin{abstract}
General movement assessment (GMA) of infant movement videos (IMVs) is an effective method for the early detection of cerebral palsy (CP) in infants.
Automated body parsing is a crucial step towards computer-aided GMA, in which infant body parts are segmented and tracked over time for movement analysis. However, acquiring fully annotated data for video-based body parsing is particularly expensive due to the large number of frames in IMVs.
In this paper, we propose a semi-supervised body parsing model, termed \textit{SiamParseNet} (SPN), to jointly learn single frame body parsing and label propagation between frames in a semi-supervised fashion.
The Siamese-structured SPN consists of a shared feature encoder, followed by two separate branches: one for intra-frame body parts segmentation, and one for inter-frame label propagation. 
The two branches are trained jointly, taking pairs of frames from the same videos as their input.
An adaptive training process is proposed that alternates training modes between using input pairs of only labeled frames and using inputs of both labeled and unlabeled frames.
During testing, we employ a multi-source inference mechanism, where the final result for a test frame is either obtained via the segmentation branch or via propagation from a nearby \textit{key} frame.
We conduct extensive experiments on a partially-labeled IMV dataset where SPN outperforms all prior arts, demonstrating the effectiveness of our proposed method.
\end{abstract}

\section{Introduction}
Early detection of cerebral palsy (CP), a group of neurodevelopmental disorders which permanently affect body movement and muscle coordination~\cite{richards2013cerebral}, enables early rehabilitative interventions for high-risk infants. One effective CP detection method is general movement assessment (GMA)~\cite{adde2007general}, where qualified clinicians observe the movement of infant body parts from captured videos and provide evaluation scores. For computer-assisted GMA, accurate segmentation of infant body parts is a crucial step.
Among different sensing modalities~\cite{hesse2018learning,Marcroft14}, RGB camera-based approaches~\cite{Adde09,zhang2019online} have numerous advantages, such as easy access to the capturing device and better recording of spontaneous infant movements. Therefore, in this paper, we focus on parsing body parts in infant movement videos (IMVs) captured by RGB cameras.

\begin{figure}[t]
    \centering
    \includegraphics[width=0.99\linewidth]{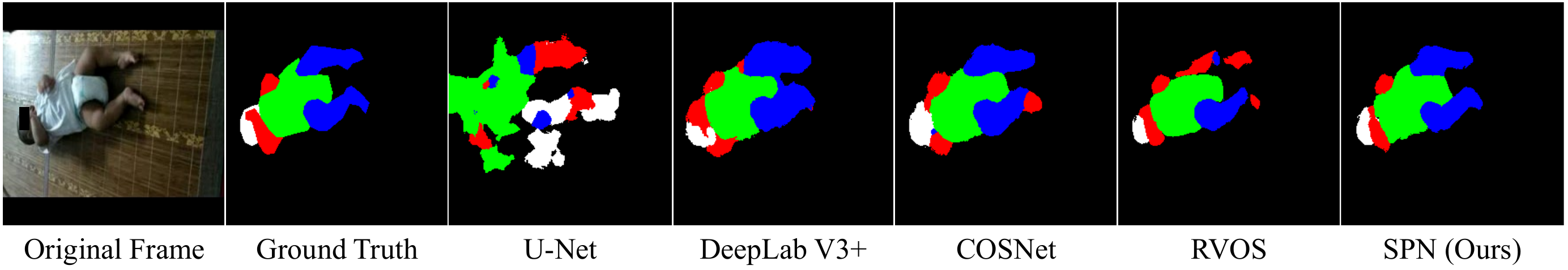}
    \caption{Comparison between different body parsing methods on IMVs.}
    \label{fig:sota}
\end{figure}

Parsing infant bodies in IMVs is closely related to the video object segmentation task (VOS), which has been widely explored in natural scenes. However, challenges arise when directly applying existing VOS methods to infant body parsing, such as frequent occlusion among body parts due to infant movements.
Among current state-of-the-art methods, Lu~\textit{et al.}~\cite{lu2019see} introduced a CO-attention Siamese Network (COSNet) to capture the global-occurrence consistency of primary objects among video frames for VOS. 
Ventura~\textit{et al.}~\cite{ventura2019rvos} proposed a recurrent network RVOS to integrate spatial and temporal domains for VOS by employing uni-directional convLSTMs~\cite{xingjian2015convolutional}. 
Zhu~\textit{et al.}~\cite{zhu2017deep} proposed Deep Feature Flow to jointly learn image recognition and flow networks, which first runs the convolutional sub-network on key frames and then propagates their deep feature maps to other frames via optical flow.
Also using optical flow, in the medical video field, Jin~\textit{et al.}~\cite{jin2019incorporating} proposed MF-TAPNet for instrument segmentation in minimally invasive surgery videos, which incorporates motion flow-based temporal prior with an attention pyramid network. 
Optical flow-based methods aim to find point-to-point correspondences between frames and have achieved promising results. However, for infant videos with frequent occlusions, it can be challenging to locate corresponding points on occluded body parts. 
Moreover, few previous methods have investigated the semi-supervised training setting. As the majority of IMV frames are unlabeled due to the high cost of annotation, semi-supervised methods have great potential in IMV body parsing and deserve further research.

In this paper, we aim to address the aforementioned challenges by proposing \textit{SiamParseNet} (SPN), a semi-supervised framework based on a siamese structure \cite{bertinetto2016fully} that automatically segments infant body parts in IMVs.
The SPN consists of one shared feature encoder $\Delta_\text{enc}$, one intra-frame body part segmentation branch $\Gamma_\text{seg}$, and one inter-frame label propagation branch
$\Gamma_\text{prop}$. 
$\Gamma_\text{prop}$ is designed to take into consideration multiple possible correspondences when calculating the label probabilities for one point to mitigate the occlusion issue.
To encourage consistent outputs from $\Gamma_\text{seg}$ and $\Gamma_\text{prop}$, we further introduce a consistency loss between their outputs as a mutual regularization. During testing, we propose a multi-source inference (MSI) scheme to further improve body parsing performance. MSI splits the whole video into multiple short clips and selects a \textit{key} frame to represent each clip. Since propagations from key frames are often more accurate, MSI utilizes the trained segmentation branch to generate results for key frames and the trained propagation branch to generate results for other non-key frames.
Our proposed methods are validated on an IMV dataset~\cite{zhang2019online} and have shown noticeably better performances when compared with several other state-of-the-art image/video segmentation methods~\cite{chen2018encoder,lu2019see,ronneberger2015u,ventura2019rvos}. An example of comparing the result of SPN with other methods is shown in Fig.~\ref{fig:sota}. We also conduct various ablation studies to show the importance of our proposed joint training, consistency loss, adaptive semi-supervised training process, and multi-source inference.
To the best of our knowledge, our work is the first attempt towards solving the IMV infant body parsing problem under the semi-supervised learning setting. 

\section{Methodology} \label{sec:Method}
\begin{figure}[t]
    \centering
    \includegraphics[width=\linewidth]{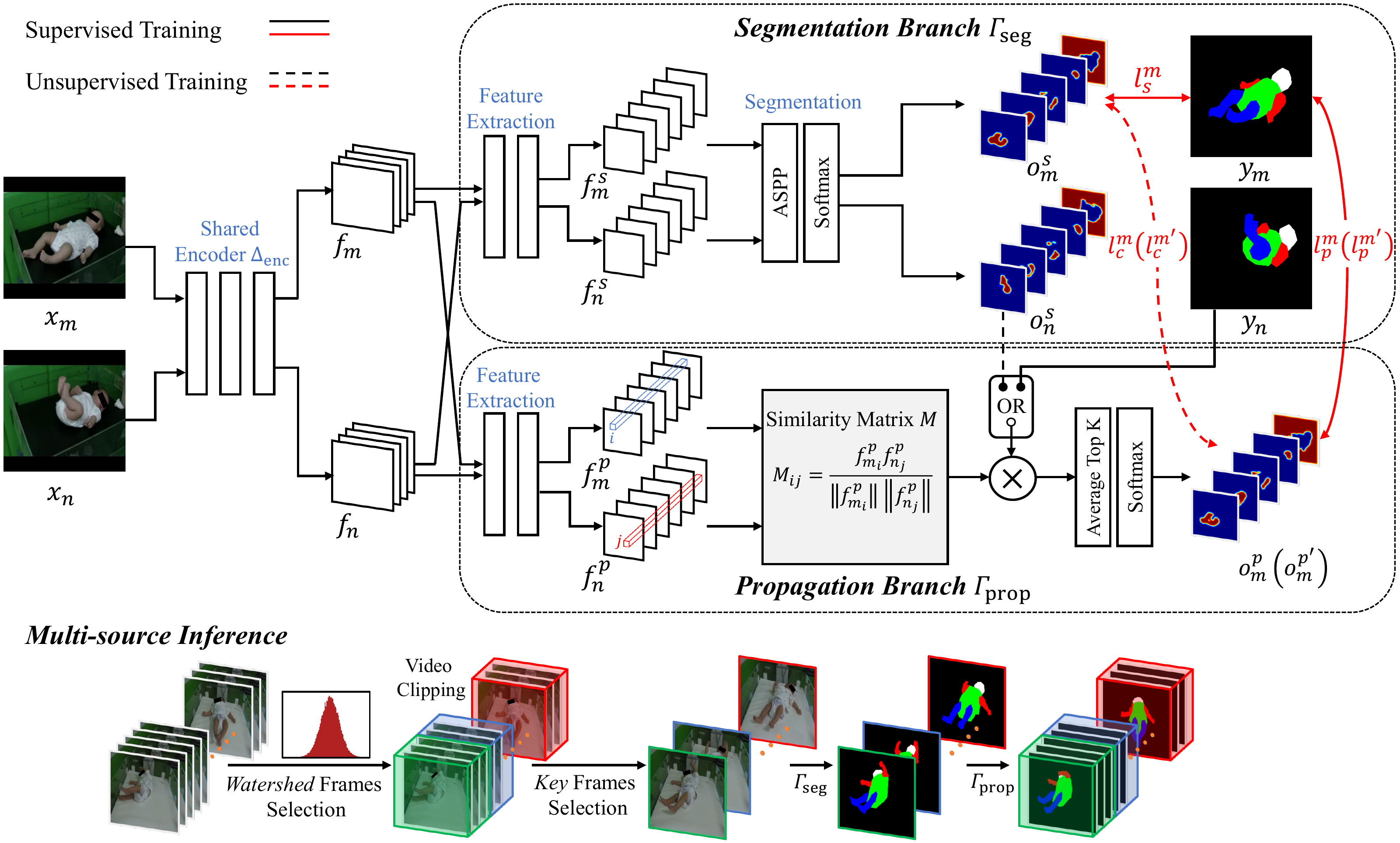}
    \caption{The overall framework of SiamParseNet (SPN). Top: Semi-supervised Training framework. Bottom: Multi-source Inference for generating test results. }
    \label{fig:dualnet}
\end{figure}

\noindent Fig.~\ref{fig:dualnet} shows the overall framework of our proposed SiamParseNet (SPN). 
The training of SPN takes as input a pair of frames from the same video, and the inherent siamese structure of SPN enables its training with any pair of frames regardless of the availability of annotation. The flexibility of the siamese structure of SPN significantly augments the number of available training examples as well as maximizes the utilization of partially labeled data. More specifically, we consider three different training modes according to the availability of labels for a paired input: the fully supervised mode which takes as input a pair of frames that are both annotated, the semi-supervised mode which has one frame in the input pair annotated, and the unsupervised mode which takes a pair of input frames that are both unannotated. During training, we adopt an adaptive alternative training (AAT) process that relies more on the supervised mode at the early stages and then gradually incorporate more semi-supervised and unsupervised training at later stages. During testing, a multi-source inference mechanism is utilized to achieve robust segmentation taking advantage of both the trained segmentation branch $\Gamma_\text{seg}$ and propagation branch $\Gamma_\text{prop}$. 
Next, we explain the SPN semi-supervised training framework, the adaptive alternative training process, and the multi-source inference procedure in more details.

\subsection{Semi-supervised Learning}
 Given a pair of input frames $x_m$ and $x_{n}$, we first employ the shared encoder $\Delta_\text{enc}$ to extract their visual feature maps $f_m$ and $f_{n}$. We then feed this pair of feature maps to branch $\Gamma_\text{seg}$ and $\Gamma_\text{prop}$.
$\Gamma_\text{seg}$ further processes $f_m$ and $f_{n}$ as $f^s_{m}$ and $f^s_{n}$ and generates segmentation probability maps $o^s_{m}$ and $o^{s}_{n}$ with a segmentation network.
$\Gamma_\text{prop}$ encodes $f_m$ and $f_{n}$ as $f^p_{m}$ and $f^p_{n}$ to
calculate a similarity matrix $M$ in feature space. Using $M$, $\Gamma_\text{prop}$ can generate the final segmentation maps $o^{p}_{n}$ and $o^{p}_{m}$ through different paths, depending on the availability of ground truth labels for $x_m$ and $x_{n}$.

\noindent\textbf{Case 1:} When both frames $x_m$ and $x_{n}$ are annotated with ground truth label maps $y_m$ and $y_n$, we have the fully supervised training mode. In this mode, $\Gamma_\text{prop}$ takes the ground truth segmentation map from one frame and propagates it to another frame (solid line in Fig.~\ref{fig:dualnet}). The overall loss $l_\text{sup}$ is calculated as:
\begin{equation}
    \label{eq:l_sup}
    l_\text{sup} = l^{m}_{s} + l^{n}_{s} + l^{m}_{p} + l^{n}_{p} 
    + \lambda\left(l^{m}_{c}+l^{n}_{c}\right)\enspace,
\end{equation}
\noindent where all losses are cross-entropy losses between one-hot vectors. More specifically, $l^{m}_{s}$ and $l^{n}_{s}$ are segmentation losses of $\Gamma_\text{seg}$ between $o^s_m$ and $y_m$, $o^s_n$ and $y_n$, respectively. $l^{m}_{p}$ and $l^{n}_{p}$ are losses of $\Gamma_\text{prop}$ between $o^{p}_m$ and $y_m$, $o^{p}_n$ and $y_n$.
The consistency loss $l_{c}$ measures the degree of overlapping between outputs of the two branches $\Gamma_\text{seg}$ and $\Gamma_\text{prop}$, 
where $\lambda$ is a scaling factor to ensure $l_{c}$ have roughly the same magnitude as $l_{s}$ and $l_{p}$.

\noindent\textbf{Case 2:} When none of the input frames are labeled, which is the unsupervised training mode, $\Gamma_\text{prop}$ propagates the segmentation map of one frame generated by $\Gamma_\text{seg}$ to another (dotted line in Fig.~\ref{fig:dualnet}). More specifically, $\Gamma_\text{prop}$ utilizes the outputs of $\Gamma_\text{seg}$, $o^s_m$ and $o^s_n$, and transforms them to obtain its probabilistic maps $o^{p'}_{n}$ and $o^{p'}_{m}$.
Since no ground truth label is available, the loss is calculated as the consistency loss between the two branches:
\begin{equation}
    \label{eq:l_un}
    l_\text{un}= \lambda\left(l^{m'}_{c}+l^{n'}_{c}\right)\enspace.
\end{equation}

\noindent\textbf{Case 3:} If only one input frame is annotated, we have the semi-supervised training mode.  Without loss of generality, let us  assume that $y_m$ is available and $y_n$ is not. Then, $\Gamma_\text{prop}$ propagates from $\Gamma_\text{seg}$'s output $o^s_{n}$ instead of $y_n$ to generate the probability map $o^{p'}_{m}$, as the dotted line in Fig.~\ref{fig:dualnet} shows.
We compute the losses $l^{m'}_{p}$ and $l^{m'}_{c}$, which measure the loss between $o^{p'}_{m}$ and $y_m$, $o^{p'}_{m}$ and $o^s_m$, respectively.
We also calculate $l^{m}_{s}$ and $l^{n}_{c}$, which measure the loss between $o^s_m$ and $y_m$, $o^p_n$ and $o^s_n$, respectively. 
Thus, the overall loss is:
\begin{equation}
    \label{eq:l_semi}
    l_\text{semi}= l^{m}_{s} + l^{m'}_{p} + \lambda\left(l^{m'}_{c}+l^{n}_{c}\right)\enspace.
\end{equation}

As a general framework, SPN can employ various networks as its backbone. Without loss of generality, we follow DeepLab~\cite{chen2017deeplab,chen2018encoder} and choose the first three blocks of ResNet101~\cite{he2016deep} as encoder $\Delta_\text{enc}$. 
For branch $\Gamma_\text{seg}$, we employ the $4^{th}$ block of ResNet101 to further extract features with the ASPP~\cite{chen2017deeplab} module. Branch $\Gamma_\text{prop}$ also utilizes the $4^{th}$ block of ResNet101 for feature representation. Note that the two branches do not share the same weights and the weights are updated separately during training.

To propagate a given source segmentation map to a target frame, similar to~\cite{hu2018videomatch}, we first calculate the cosine similarity matrix $M$ of $f^{p}_{m}$ and $f^{p}_n$ as 
\begin{equation}
    \label{eq:cos}
    M_{ij}= \frac{f^{p}_{m_i}\cdot f^{p}_{n_j}}{\left\|f^{p}_{m_i}\right\|\left\|f^{p}_{n_j}\right\|}\enspace,
\end{equation}
where $M_{ij}$ is the affinity value between $f^{p}_{m_i}$, point $i$ in map $f^{p}_{m}$, and $f^{p}_{n_j}$, point $j$ in map $f^{p}_n$. $\left\|\cdot\right\|$ indicates the L2 norm. Then, given the source segmentation map $\hat{y}_n$ (either ground truth $y_n$ or $\Gamma_\text{seg}$ generated $o^s_n$), and the similarity matrix $M$, $\Gamma_\text{prop}$ produces $o^{p}_{m_i}$, point $i$ in output map $o^{p}_m$ as
\begin{equation}
    \label{eq:sim}
    o^{p}_{m_i} = \text{softmax}\left(\frac{1}{K}\sum\nolimits_{j\in \text{Top}(M_i, K)} M_{ij}\hat{y}_{n_j}\right)\enspace,
\end{equation}
where $\text{Top}(M_i, K)$ contains the indices of the top $K$ most similar scores in the $i^{th}$ row of $M$. 
Since $\Gamma_\text{prop}$ considers multiple correspondences for a point rather than one-to-one point correspondences as in optical flow \cite{jin2019incorporating,meister2018unflow,zhu2017deep}, SPN can naturally better handle occlusions in IMVs than optical flow based methods. 

\subsection{Adaptive Alternative Training}
As mentioned briefly in the beginning of Section \ref{sec:Method}, during the training of SPN, we adopt an adaptive alternative training (AAT) process to alternatively use different training modes to achieve optimal performance.  Intuitively, SPN should rely more on the supervised mode at early stages and then gradually incorporate more semi-supervised and unsupervised training at later stages of training. To dynamically adjust the proportion of different training modes, we propose AAT to automatically sample training data among the three cases. Assume that the probabilities of selecting case 1, case 2, and case 3 for any iteration/step of training are $p_1$, $p_2$, and $p_3$, respectively. Considering case 2 and case 3 both involve utilizing unlabeled frames, we set $p_2 = p_3 = \frac{1-p_1}{2}$. 
Thus we only need to control the probability of choosing case 1 training and the other two cases are automatically determined. For AAT, we use an annealing temperature $t$ to gradually reduce $p_1$ as training continues, where $p_1$ is computed as
\begin{equation}
    \label{eq:ada}
        p_1 = \text{max}\left(1-
        \left(
        1-\mathbb{P}_1
        \right)
        \left(\frac{i}{i_\text{max}}\right)^t, \mathbb{P}_1\right)\enspace,
\end{equation}
where $i$ is the training step, $\mathbb{P}_1$ is the pre-defined lower bound probability of $p_1$, $i_\text{max}$ is the maximum number of steps of using AAT.

\subsection{Multi-source Inference for Testing}
\label{sec:infer}
To further mitigate the occlusion issue during testing, we exploit the capabilities of both trained branches in SPN by a multi-source inference (MSI) mechanism, as illustrated in the bottom part of Fig.~\ref{fig:dualnet}. While the propagation branch could generate more desirable results than the segmentation branch, an appropriate source frame needs to be chosen to alleviate the occlusion issue during propagation. To this end, we propose to choose \textit{key} frames as source frames for propagation.
For each testing IMV, we calculate the pixel differences between consecutive frames. The differences are modeled by a Gaussian distribution, and its $\alpha$-th percentile is chosen as the threshold to sample a series of \textit{watershed} frames, whose pixel differences are higher than the threshold. We then segment the video into multiple clips delimited by watershed frames, so that infant poses and appearances are similar within the same clip.
Then, we choose the middle frame of each clip as the \textit{key} frame for that clip. The intuition behind the choice is that middle frames have the least cumulative temporal distance from other frames~\cite{griffin2019bubblenets} and thus can better represent the clip that it is in.  
During inference, $\Gamma_\text{seg}$ first segments the selected key frames to provide propagation sources. Then, for other non-key frames within a clip, $\Gamma_\text{prop}$ takes the corresponding key frame's segmentation output and propagates it to all other non-key frames in the clip.
By splitting a long video into short video clips and using key frames to provide local context and source of propagation within each short clip, the proposed MSI can effectively mitigate the occlusion problem in IMVs. 

\section{Experiments}
We conduct extensive experiments on an IMV dataset collected from a GMA platform \cite{zhang2019online}. 20 videos with infants' ages ranging from 0 to 6 months are recorded by either medical staff in hospitals or parents at home. The original videos are very long and they are downsampled every 2 to 5 frames to remove some redundant frames. All frames are resized to the resolution of $256\times 256$ and some of the frames are annotated with five categories: background, head, arm, torso, and leg. 
This challenging dataset covers large varieties of video content, including different infant poses, body appearance, diverse background, and viewpoint changes.
We randomly divide the dataset into 15 training videos and 5 testing videos, resulting in $1,267$ labeled frames and $21,154$ unlabeled frames in the training set, and 333 labeled frames and $7,246$ unlabeled frames in the testing set. We evaluate all methods with Dice coefficient on labeled testing frames. We represent each pixel label in the segmentation map as a one-hot vector, where background pixels are ignored when calculating Dice to focus on the body parts. Dice scores of each labeled testing frame are averaged to get the final mean Dice. 

\begin{table}[t]
    \begin{minipage}[c]{0.28\linewidth}
    \centering
    \resizebox{\linewidth}{!}{
    \begin{tabular}{l|cccc|c}
        \hline
        Methods                 & Dice  \\
        \hline
        U-Net \cite{ronneberger2015u}       & 46.12          \\
        DeepLab V3+ \cite{chen2018encoder} & 73.32          \\
        COSNet \cite{lu2019see}      & 79.22          \\
        RVOS \cite{ventura2019rvos}        & \textbf{81.98}          \\
        \hline
        SPN w/o SSL          & {82.31} \\
        SPN & \textbf{83.43} \\
        \hline
    \end{tabular}
    }
    \subcaption{Comparison between SPN and current state-of-the-art methods.}
    \label{tab:sota}
    \end{minipage}
    \begin{minipage}[c]{0.31\linewidth}
    \centering
    \resizebox{0.80\linewidth}{!}{
    \begin{tabular}{l|c}
        \hline
        Methods                     & Dice  \\
        \hline
        Single-$\Gamma_\text{seg}$  & 71.82 \\
        SPN-$\Gamma_\text{seg}$ & \textbf{81.33} \\
        \hline
        SPN w/o $l_c$ & 81.38          \\
        SPN w/o SSL            & \textbf{82.31} \\
        \hline
        SPN $t=1.4$   & 82.88          \\
        SPN $t=0.9$   & 82.79          \\
        SPN $t=0.4$   & \textbf{83.43} \\
        \hline
    \end{tabular}
    }
    \subcaption{Ablation study of SPN.}
    \label{tab:aba}
    \end{minipage}
    \begin{minipage}[c]{0.33\linewidth}
    \centering
    \resizebox{\linewidth}{!}{
    \begin{tabular}{l|c|c|c}
    \hline
        Methods          & 0/15 & 8/15  & 10/15  \\ \hline
        SPN w/o SSL      & 82.31 & 77.83 & 72.93 \\
        SPN & 83.43 & 80.50 & 78.56 \\ \hline
        Dice Gain             & 1.12  & 2.67  & \textbf{5.63}  \\ \hline
    \end{tabular}
    }
    \subcaption{SPN without and with SSL trained using different numbers of completely unlabeled videos, out of the 15 training videos.}
    \label{tab:semi}
    \end{minipage}
    \caption{Quantitative comparison of different methods.} 
    \label{tab:all}
\end{table}

\subsection{Implementation Details}
To accelerate training and reduce overfitting, similar to~\cite{chen2017deeplab,chen2018encoder,lu2019see,ventura2019rvos}, we initialize the network parameters using weights from DeepLab V3+~\cite{chen2018encoder} pretrained on the COCO dataset~\cite{lin2014microsoft} for shared encoder $\Delta_\text{enc}$, segmentation branch $\Gamma_\text{seg}$ and propagation branch $\Gamma_\text{prop}$. For $\Gamma_\text{prop}$, we set $K$ to be $20$, after grid searching over the values of $K$ from 5 to 20 with interval 5. The scaling factor $\lambda$ in Eq.~\ref{eq:l_sup} is set to be $10^{-6}$.
To build the training set for SPN that consists of pairs of frames from the same videos, we collect input pairs for the three training modes separately. For training case 1 (\textit{i.e.}, fully supervised mode), we take two randomly selected labeled frames and repeat the operation    
until $10,000$ image pairs are selected from $1,267$ labeled training frames, with $20,000$ images in total.  For case 2 (\textit{i.e.}, unsupervised mode), we randomly select two unlabeled frames and repeat until we have $10,000$ image pairs. For case 3 (\textit{i.e.}, semi-supervised mode), we randomly choose one labeled frame and one unlabeled frame in the same video and repeat until we have $10,000$ image pairs. Then, the sampling of training input pairs from the built set follows Eq.~\ref{eq:ada}, where we set $\mathbb{P}_1$ to be $1/3$ and $i_\text{max}$ to be 20 epochs. Unless otherwise specified, $t$ is set to be $0.4$ in all experiments.  We use the SGD optimizer with momentum $0.9$. We set the initial learning rate to be $2.5\times 10^{-4}$ and adopt the poly learning rate policy~\cite{chen2017deeplab} with power of $0.9$. The training batch size is set to be $20$. Traditional data augmentation techniques, such as color jittering, rotation, flipping, are also applied. We terminate the training until the pixel-level accuracy of both branches remains mostly unchanged for 2 epochs. During inference, we set $\alpha$ in the key frame selection algorithm to be $0.98$.

\subsection{Result Analysis}
We compare our proposed SPN with current state-of-the-art methods, including single frame based U-Net~\cite{ronneberger2015u} and DeepLab V3+~\cite{chen2018encoder}, and video based COSNet~\cite{lu2019see} and RVOS~\cite{ventura2019rvos} in Fig.~\ref{fig:sota} and Table~\ref{tab:all}(\subref{tab:sota}). For fair comparison, except U-Net, all methods employ pretrained models from ImageNet \cite{deng2009imagenet} or COCO dataset \cite{lin2014microsoft}. DenseCRF~\cite{krahenbuhl2011efficient} is also adopted as post-processing for all methods. From Fig.~\ref{fig:sota}, one can observe that SPN clearly handles occlusion better than other methods and shows better qualitative results.
As Table~\ref{tab:all}(\subref{tab:sota}) shows, SPN without semi-supervised learning (\textit{i.e.}, fully supervised mode only) and the full SPN model with SSL have achieved substantially better quantitative performance when compared with previous state-of-the-art methods.

To validate the effect of joint training of $\Gamma_\text{seg}$ and $\Gamma_\text{prop}$,
we compare two variants of SPN under fully supervised training mode: [Single-$\Gamma_\text{seg}$] which is trained using only segmentation branch and [SPN-$\Gamma_\text{seg}$] which is trained jointly but only $\Gamma_\text{seg}$ is used for all testing frames. Since $\Gamma_\text{prop}$ is not available, multi-source inference  is not used during test.
As Table~\ref{tab:all}(\subref{tab:aba}) shows, compared with [Single-$\Gamma_\text{seg}$], joint training of SPN greatly boosts the mean dice of $\Gamma_\text{seg}$ by over $9\%$, which can be contributed to the siamese structure, the shared encoder, the consistency loss, among others.
To further validate the effect of the consistency loss, we also compare two variants of the SPN model: [SPN w/o $l_c$] which is SPN trained without using consistency loss and SSL, and [SPN w/o SSL] which is SPN trained with consistency loss but without SSL. For both variants, MSI is used for testing.
As shown in Table~\ref{tab:all}(\subref{tab:aba}), [SPN w/o SSL] gives better mean Dice than [SPN w/o $l_c$], which demonstrates the usefulness of the consistency loss, even in the fully supervised setting.
In addition, by comparing the results of [SPN-$\Gamma_\text{seg}$] and [SPN w/o SSL], one can see the effectiveness of MSI since the only difference between those two models is the multi-source inference.

\begin{figure}[t]
    \centering
    \includegraphics[width=0.75\linewidth]{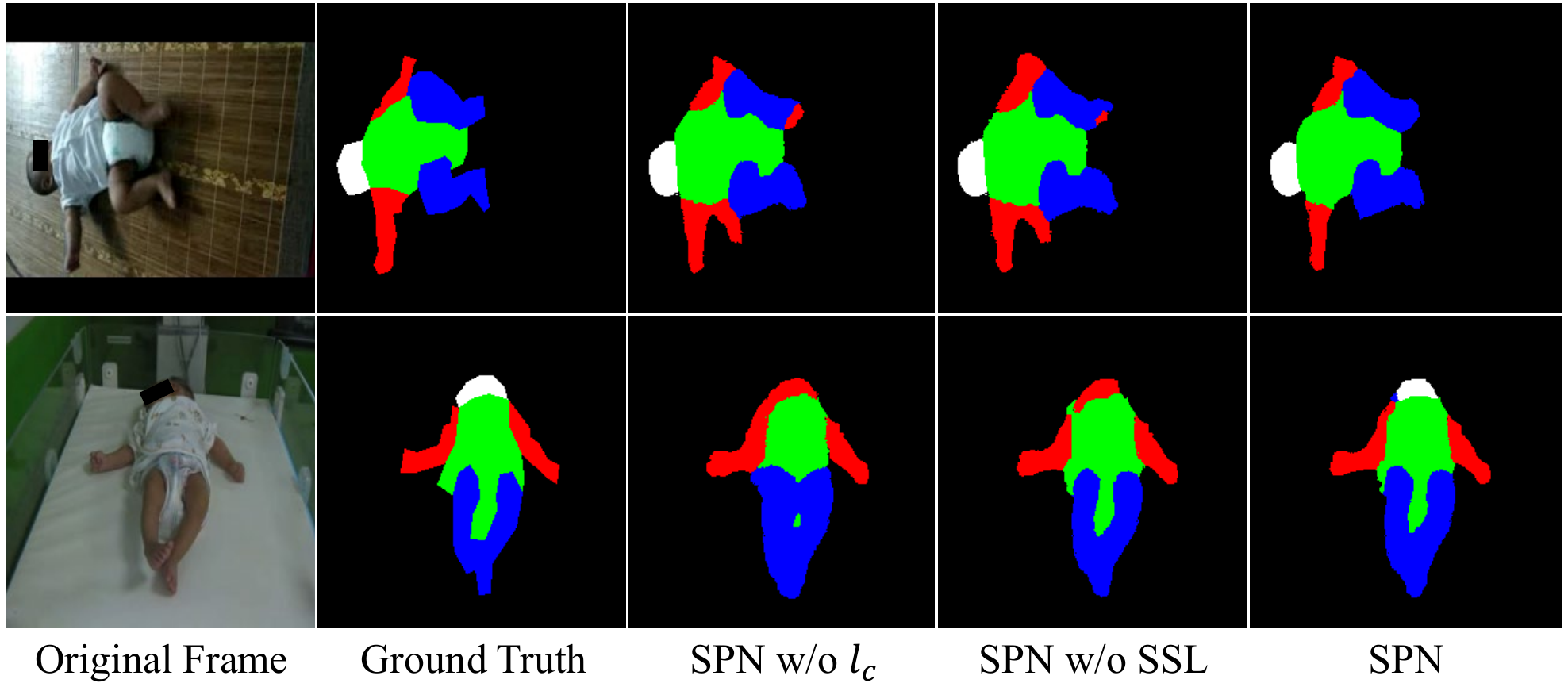}
    \caption{Qualitative comparison of ablation study.}
    \label{fig:visual}
\end{figure}

For full SPN model with SSL, we experiment with different values of the annealing temperature $t$ in AAT (Eq.~\ref{eq:ada}). From Table~\ref{tab:all}(\subref{tab:aba}), one can observe that [SPN w/ SSL] outperforms all other variants including the one without SSL, and the best model comes with $t=0.4$, which we set as the default value in all other experiments. Qualitatively, from Fig.~\ref{fig:visual}, one can see that different from other variants of SPN, our full SPN model avoids mis-segmentation of shadows and occluded head region, giving the best segmentation performance.

To further demonstrate the power of semi-supervised learning in our proposed SPN, we experiment with a \textit{video-level} SSL setting: we randomly choose a certain number of training videos and remove \textbf{ALL} annotations from those videos. This setting is more stringent than the one used in \cite{jin2019incorporating}, which showed promising results with a \textit{frame-level} SSL setting: they removed labels of some frames in each video but all training video are still partially labeled. For our experiment using the \textit{video-level} SSL, we compare the performances of the [SPN w/o SSL] and the full SPN models when only keeping labeled frames in 7 training videos (and using 8 training videos without labels), and when only keeping labels in 5 videos (and using 10 without labels).
As shown in Table~\ref{tab:all}(\subref{tab:semi}), with fewer annotated videos, the full SPN with SSL clearly shows significant performance gains (up to $5.63\%$) than [SPN w/o SSL]. Such results indicate the potential of SPN under various semi-supervised learning scenarios.

\section{Conclusions}
In this paper, we propose SiamParseNet, a novel semi-supervised framework for joint learning of body parsing and label propagation in IMVs toward computer-assisted GMA. Our proposed SPN exploits a large number of unlabeled frames in IMVs via adaptive alternative training of different training modes and shows superior performance under various semi-supervised training settings. Combined with multi-source inference for testing, SPN not only has great potential in infant body parsing but can also be easily adapted to other video based segmentation tasks such as instrument segmentation in surgical videos.

\bibliographystyle{splncs04}

\begin{thebibliography}{10}
\providecommand{\url}[1]{\texttt{#1}}
\providecommand{\urlprefix}{URL }
\providecommand{\doi}[1]{https://doi.org/#1}

\bibitem{Adde09}
Adde, L., Helbostad, J., Jensenius, A., G.Taraldsen, Støen, R.: Using
  computer-based video analysis in the study of fidgety movements. Early Human
  Development  \textbf{85} (2009)

\bibitem{adde2007general}
Adde, L., Rygg, M., Lossius, K., {\O}berg, G.K., St{\o}en, R.: General movement
  assessment: predicting cerebral palsy in clinical practise. Early Human
  Development  \textbf{83} (2007)

\bibitem{bertinetto2016fully}
Bertinetto, L., Valmadre, J., Henriques, J.F., Vedaldi, A., Torr, P.H.:
  Fully-convolutional siamese networks for object tracking. In: Proceedings of
  the European Conference on Computer Vision (ECCV). pp. 850--865. Springer
  (2016)

\bibitem{chen2017deeplab}
Chen, L.C., Papandreou, G., Kokkinos, I., Murphy, K., Yuille, A.L.: Deeplab:
  Semantic image segmentation with deep convolutional nets, atrous convolution,
  and fully connected crfs. IEEE transactions on pattern analysis and machine
  intelligence  \textbf{40}(4),  834--848 (2017)

\bibitem{chen2018encoder}
Chen, L.C., Zhu, Y., Papandreou, G., Schroff, F., Adam, H.: Encoder-decoder
  with atrous separable convolution for semantic image segmentation. In:
  Proceedings of the European conference on computer vision (ECCV). pp.
  801--818. Springer (2018)

\bibitem{deng2009imagenet}
Deng, J., Dong, W., Socher, R., Li, L.J., Li, K., Fei-Fei, L.: Imagenet: A
  large-scale hierarchical image database. In: Proceedings of the IEEE
  Conference on Computer Vision and Pattern Recognition (CVPR). pp. 248--255
  (2009)

\bibitem{griffin2019bubblenets}
Griffin, B.A., Corso, J.J.: Bubblenets: Learning to select the guidance frame
  in video object segmentation by deep sorting frames. In: Proceedings of the
  IEEE Conference on Computer Vision and Pattern Recognition (CVPR). pp.
  8914--8923 (2019)

\bibitem{he2016deep}
He, K., Zhang, X., Ren, S., Sun, J.: Deep residual learning for image
  recognition. In: Proceedings of the IEEE Conference on Computer Vision and
  Pattern Recognition (CVPR). pp. 770--778 (2016)

\bibitem{hesse2018learning}
Hesse, N., Pujades, S., Romero, J., Black, M.J., Bodensteiner, C., Arens, M.,
  Hofmann, U.G., Tacke, U., Hadders-Algra, M., Weinberger, R., et~al.: Learning
  an infant body model from rgb-d data for accurate full body motion analysis.
  In: International Conference on Medical Image Computing and Computer-Assisted
  Intervention (MICCAI). pp. 792--800. Springer (2018)

\bibitem{hu2018videomatch}
Hu, Y.T., Huang, J.B., Schwing, A.G.: Videomatch: Matching based video object
  segmentation. In: Proceedings of the European Conference on Computer Vision
  (ECCV). pp. 54--70. Springer (2018)

\bibitem{jin2019incorporating}
Jin, Y., Cheng, K., Dou, Q., Heng, P.A.: Incorporating temporal prior from
  motion flow for instrument segmentation in minimally invasive surgery video.
  In: International Conference on Medical Image Computing and Computer-Assisted
  Intervention (MICCAI). pp. 440--448. Springer (2019)

\bibitem{krahenbuhl2011efficient}
Kr{\"a}henb{\"u}hl, P., Koltun, V.: Efficient inference in fully connected crfs
  with gaussian edge potentials. In: Advances in neural information processing
  systems. pp. 109--117 (2011)

\bibitem{lin2014microsoft}
Lin, T.Y., Maire, M., Belongie, S., Hays, J., Perona, P., Ramanan, D.,
  Doll{\'a}r, P., Zitnick, C.L.: Microsoft coco: Common objects in context. In:
  Proceedings of the European Conference on Computer Vision (ECCV). pp.
  740--755. Springer (2014)

\bibitem{lu2019see}
Lu, X., Wang, W., Ma, C., Shen, J., Shao, L., Porikli, F.: See more, know more:
  Unsupervised video object segmentation with co-attention siamese networks.
  In: Proceedings of the IEEE Conference on Computer Vision and Pattern
  Recognition (CVPR). pp. 3623--3632 (2019)

\bibitem{Marcroft14}
Marcroft, C., Khan, A., Embleton, N.D., Trenell, M., Plötz, T.: Movement
  recognition technology as a method of assessing spontaneous general movements
  in high risk infants. Frontiers in neurology  \textbf{5} (2014)

\bibitem{meister2018unflow}
Meister, S., Hur, J., Roth, S.: Unflow: Unsupervised learning of optical flow
  with a bidirectional census loss. In: Thirty-Second AAAI Conference on
  Artificial Intelligence (2018)

\bibitem{richards2013cerebral}
Richards, C.L., Malouin, F.: Cerebral palsy: definition, assessment and
  rehabilitation. In: Handbook of clinical neurology, vol.~111, pp. 183--195.
  Elsevier (2013)

\bibitem{ronneberger2015u}
Ronneberger, O., Fischer, P., Brox, T.: U-net: Convolutional networks for
  biomedical image segmentation. In: International Conference on Medical image
  computing and computer-assisted intervention (MICCAI). pp. 234--241. Springer
  (2015)

\bibitem{ventura2019rvos}
Ventura, C., Bellver, M., Girbau, A., Salvador, A., Marques, F., Giro-i Nieto,
  X.: Rvos: End-to-end recurrent network for video object segmentation. In:
  Proceedings of the IEEE Conference on Computer Vision and Pattern Recognition
  (CVPR). pp. 5277--5286 (2019)

\bibitem{xingjian2015convolutional}
Xingjian, S., Chen, Z., Wang, H., Yeung, D.Y., Wong, W.K., Woo, W.c.:
  Convolutional lstm network: A machine learning approach for precipitation
  nowcasting. In: Advances in neural information processing systems. pp.
  802--810 (2015)

\bibitem{zhang2019online}
Zhang, Q., Xue, Y., Huang, X.: Online training for body part segmentation in
  infant movement videos. In: 2019 IEEE 16th International Symposium on
  Biomedical Imaging (ISBI 2019). pp. 489--492. IEEE (2019)

\bibitem{zhu2017deep}
Zhu, X., Xiong, Y., Dai, J., Yuan, L., Wei, Y.: Deep feature flow for video
  recognition. In: Proceedings of the IEEE Conference on Computer Vision and
  Pattern Recognition (CVPR). pp. 2349--2358 (2017)

\end{thebibliography}

\end{document}